\documentclass{llncs}
\usepackage{llncsdoc}
\usepackage{times}
\usepackage{url}
\usepackage{latexsym}
\usepackage{algorithm}
\usepackage{algorithmic}
\usepackage{enumitem}
\usepackage{amsmath}
\usepackage{graphicx}
\providecommand{\shortcite}[1]{\cite{#1}}
\usepackage{wrapfig,lipsum,booktabs,multirow,subcaption}
\usepackage{xcolor,colortbl,framed,cuted}
\usepackage[hidelinks]{hyperref}
\providecommand{\overrightarrow}[1]{\vec{#1}}
\begin{document}

\title{Reading Comprehension using Entity-based Memory Networks}

\author{Xun Wang\inst{1} \inst{2} \and
Katsuhito Sudoh\inst{1} \and Masaaki Nagata\inst{1}\and \\ Tomohide Shibata\inst{2} \and  Daisuke Kawahara\inst{2}\and Sadao Kurohashi\inst{2}
}
\institute{
NTT Communication Science Labratories, Kyoto, Japan,\\
\email{wang.xun,sudoh.katsuhito,nagata.masaaki@lab.ntt.co.jp}
\and 
Kyoto University, Kyoto, Japan\\
\email{shibata,dk,kuro@i.kyoto-u.ac.jp}
}

\date{}
\maketitle

\begin{abstract}
This paper introduces a novel neural network model for question answering, the \emph{entity-based memory networks}. It enhances neural networks' ability of representing and calculating information over a long period by keeping records of entities contained in text.
The core component is a memory pool which comprises entities' states. These entities' states are continuously updated according to the input text. Questions with regard to the input text are used to search the memory pool for related entities and answers are further predicted based on the states of retrieved entities.
Compared with previous memory network models, the proposed model is capable of handling  fine-grained information and more sophisticated relations based on entities.
We formulated several different tasks as question answering problems and tested the proposed model. Experiments reported satisfying results. 
\keywords{Text Comprehension, Entity Memory Network, Question Answering}

\end{abstract}

\section{Introduction}
It has long been a major concern of the natural language processing (NLP) community to enable computers to understand text as humans do. A lot of NLP tasks have been tensely studied towards this goal such as information retrieval, semantical role labelling, textual entailment and so on. Among them, questions answering is of great importance and has been a huge challenge.
A question answering (QA) task is to predict an answer for a given question with regard to related information. 
It can be formulated as a map $f:\{related\_text, question\} \longrightarrow \{answer\}$ \cite{lehnert1978process}. 
To predict the correct answer, computers are firstly required to ``understand" the text. 

Shallow features such as bag-of-words, token frequencies and so on are unable to capture the rich information in text. Often outside knowledge is required towards better performances.
Traditional approaches heavily rely on rules or structured knowledge developed by experts or crowd sourcing \cite{riloff2000rule,poon2010machine}. Relational databases constructed from predicate argument triples also serve as a source of knowledge \cite{lin2001discovery,shen2007using}. Problems with these approaches lie in at least two aspects. Firstly the construction of structured knowledge is both time and money consuming. Secondly it is a huge challenge to design models flexible and powerful enough to learn to employ these information extracted \cite{hermann2015teaching}. Thus the progress of using machine learning for QA has been slow.

Recently the emergence of deep neural networks and distributed representations sheds light on such methods.
Representing all the features using vectors provides a unified representational form for all the necessary information. Outside knowledge learnt from large corpus can be encoded into word vectors. Information obtained locally is also represented using vectors. Deep neural network models with many layers are designed to fuse information obtained from different sources \cite{dong2015question,iyyer2014neural}.

A notable breakthrough is to employ memories in neural networks. The representative model is named the memory network \cite{weston2014memory}. The key of memory network is to store historical sentences in a memory pool. The model is trained to look for related sentences when a question comes. Then based on the related sentences, an answer is predicted for the question. Memory network remembers all sentences it has read so that it can look for useful ones when facing questions. This model and its variants have been proved useful in a series of tasks \cite{weston2014memory,sukhbaatar2015end,bordes2015large}.

One problem with memory networks is that using sentence vectors as elementary units of information makes it difficult to fully explore the information contained in text. Often is the case that in a long sentence, only part of the sentence is related to the questions.
Therefore taking the whole sentence into consideration makes it hard to focus on the information that are related to questions. Besides, learning sentence representations itself is a growing field. 

We propose to focus on entities rather than sentences.
Entities refer to anything that exist in reality or are purely hypothetical.
We assume that text can be projected to a world of entities. The key of conducting comprehension and reasoning over text is to identify its containing entities and analyze the states of these entities and the relations between them. We keep a memory pool of entities and use the input sentences to update the states of these entities. Questions are answered based on the states of related entities. The proposed model deals with fine-grained information by using entities.
The introduced model is named as \emph{entity-based memory network}.
It is tested on several datasets, including the toy bAbI dataset \cite{weston2015towards}, large movie review dataset \cite{maas-EtAl:2011:ACL-HLT2011} and the machine comprehension test dataset \cite{richardson2013mctest}. Results show we have achieved satisfying results using the entity-based memory network. The rest of the paper is organized as follows:  Section 2 reviews some previous work. Section 3 describes our approaches and elaborates the details. Section 4 presents the experiments and the analysis. Section 5 concludes the paper.

\section{Memories in Deep Neural Networks}
QA has a long history and a lot of methods have been developed to address this problem \cite{ravichandran2002learning,lin2001discovery,brill2001data}. 
Recently the development of neural models leads to series of work on question answering \cite{graves2014neural,dong2015question,iyyer2014neural}.
Among them closely related to our work is the Memory Network (MNN) \cite{weston2014memory}. The memory network contains four parts: the input module which converts sentences into vectors, the memory which keeps all sentence vectors a retrieval module and a response module. Whenever a question comes, the question is turned into a vector and the question vector is used to retrieve the memory for related sentences. The response module is used to predict an answer based on the related sentences.
The core component is the memory pool that stores all the input sentences so that they can be retrieved later to answer questions. This model contains several neural networks which are jointly optimized according to the task. Experiments on a toy dataset show that this model is able to answer simple questions according to the input text. Fig. 1(a) illustrates the memory network. 

\begin{figure}
\centering
  \begin{subfigure}[b]{0.42\textwidth}
    \includegraphics[width=\textwidth]{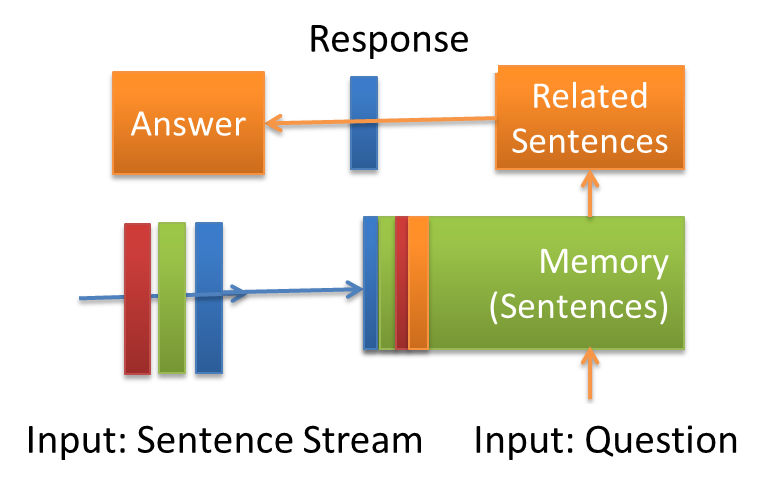}
    \caption{the Memory Network}
    \label{fig:1}
  \end{subfigure}
  \begin{subfigure}[b]{0.42\textwidth}
    \includegraphics[width=\textwidth]{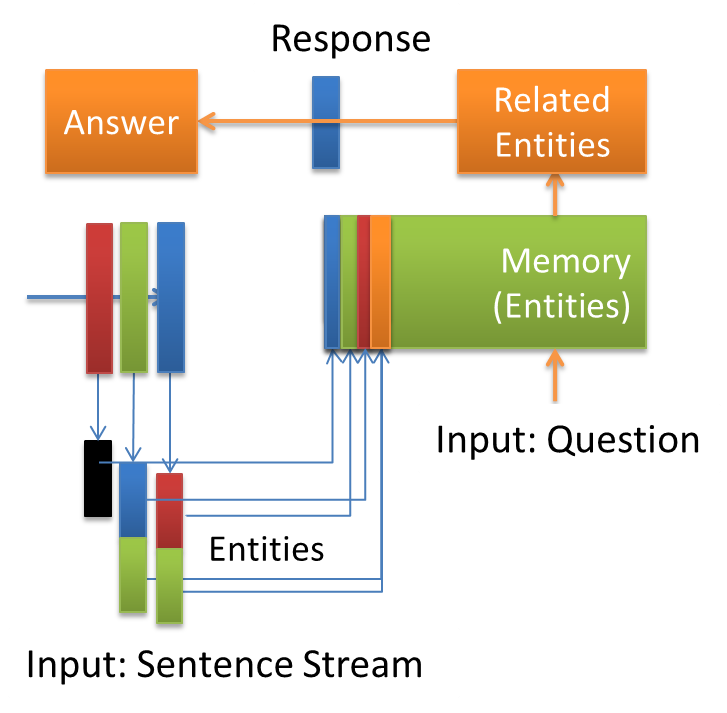}
    \caption{the Entity-based Memory Network}
    \label{fig:2}
  \end{subfigure}
  \caption{Comparison of the memory network and the proposed entity-based memory network model. Sentences are decomposed into entities and then stored in the memory for later retrieval.}
\end{figure}

Later \shortcite{kumar2015ask} propose the Dynamic Memory Network (DMNN) which introduces the attention mechanism into the memory network model. When retrieving memories, the location of the next related sentence is predicted according to the related sentences identified in the previous iterations. Using the attention mechanism, they obtain further improvements. Some other work \cite{sukhbaatar2015end,bordes2015large} propose other variants of MNN by introducing additional memory network modules. These work focuses on storing sentence vectors for later retrieval with no exceptions.
Most of them have been tested on the toy dataset bAbI \cite{weston2015towards} and are reported to have achieved satisfying results. When further tested on some practical tasks, these models also show the ability to produce results as good as existing state-of-the-art systems or even better results.
Memory networks store sentence vectors as memories and have the superiority of processing information from a large scale. Experiment results they reported on a series of tasks are concrete proofs. 

But there is also a problem with the memory networks as we have stated. Taking sentence vectors as input means that it is difficult to further analyze and take advantages of relations between smaller text units, such as entities. For example, when an entity $e_a$ of sentence $A$ interacts with another entity $e_b$ of sentence $B$, we have to take the whole sentences $A$ and $B$ into consideration rather than just focus on $e_a$ and $e_b$. This inevitably brings about noise and damages the comprehension of text. The failure of obtaining fine-grained information prevents any further improvements. In the proposed entity-based model, we focus on entities directly and avoid bringing in redundant information. 

\section{Approaches}

\subsection{Overview}
Firstly we use an example to illustrate how the model works.
Below we show a piece of text which contains 4 sentences and 2 questions. There are 7 entities in total, all of them underlined.
\begin{figure}

1) \underline{Mary} moved to the \underline{bathroom}.
2) \underline{John} went to the \underline{hallway}.
3) Where is Mary?        Bathroom.        \\
4) \underline{Daniel} went back to the \underline{hallway}.
5) \underline{Sandra} moved to the \underline{garden}.
6) Where is Daniel?      Hallway. 
\caption{An Example from bAbI, a toy dataset for question answering \cite{weston2015towards}.}
\end{figure}

This text is elaborated around the 7 entities. It describes how their states change (i.e., the change of a character's location) when the story goes on. Note that here all the entities are concrete concepts that exist in reality. It is also possible to talk about abstract concepts. 

The core of the proposed model are entities.
We take Sentence 1  ($S_1$) as input and extract the entities it contains \{Mary, bathroom\}. Vectors representing the states of these entities are initialized using some pre-learned word embeddings \{$\overrightarrow{Mary}$, $\overrightarrow{bathroom}$\} and stored in a memory pool. Meanwhile, we turn $S_1$ into a vector ($\overrightarrow{S_1}$) using an autoencoder model \footnote{Note that the sentence vector is not used to answer question directly and it is also plausible to use other models to learn sentence representation.}. Then we use the sentence vector $\overrightarrow{S_1}$ to update the entities' states \{$\overrightarrow{Mary}$, $\overrightarrow{bathroom}$\}. The goal is to reconstruct $\overrightarrow{S_1}$ solely from \{$\overrightarrow{Mary}$, $\overrightarrow{bathroom}$\}. In the same way, we process the following text ($S_2$) and its containing entities (John, hallway) until encounter a question ($S_3$).
$S_3$ is converted into a vector ($\overrightarrow{S_3}$) following the same method that processes previous input text. Then taking $\overrightarrow{S_3}$ as input, we retrieve related entities from the memory which now stores all the entities (Mary, bathroom, John, hallway) that appear before $S_3$. The related entities' states are then used to produce a feature vector. In this case, (Mary and bathroom) are related to the question and their states are used for constructing the feature vector.  Note the current states of the two entities (Mary and bathroom) are different from their initial values due to $S_1$.
Based on the feature vector, we then use another neural network model to predict the answer to $S_3$.

\begin{figure*}[!ht]
  \centering
    \includegraphics[width=\textwidth]{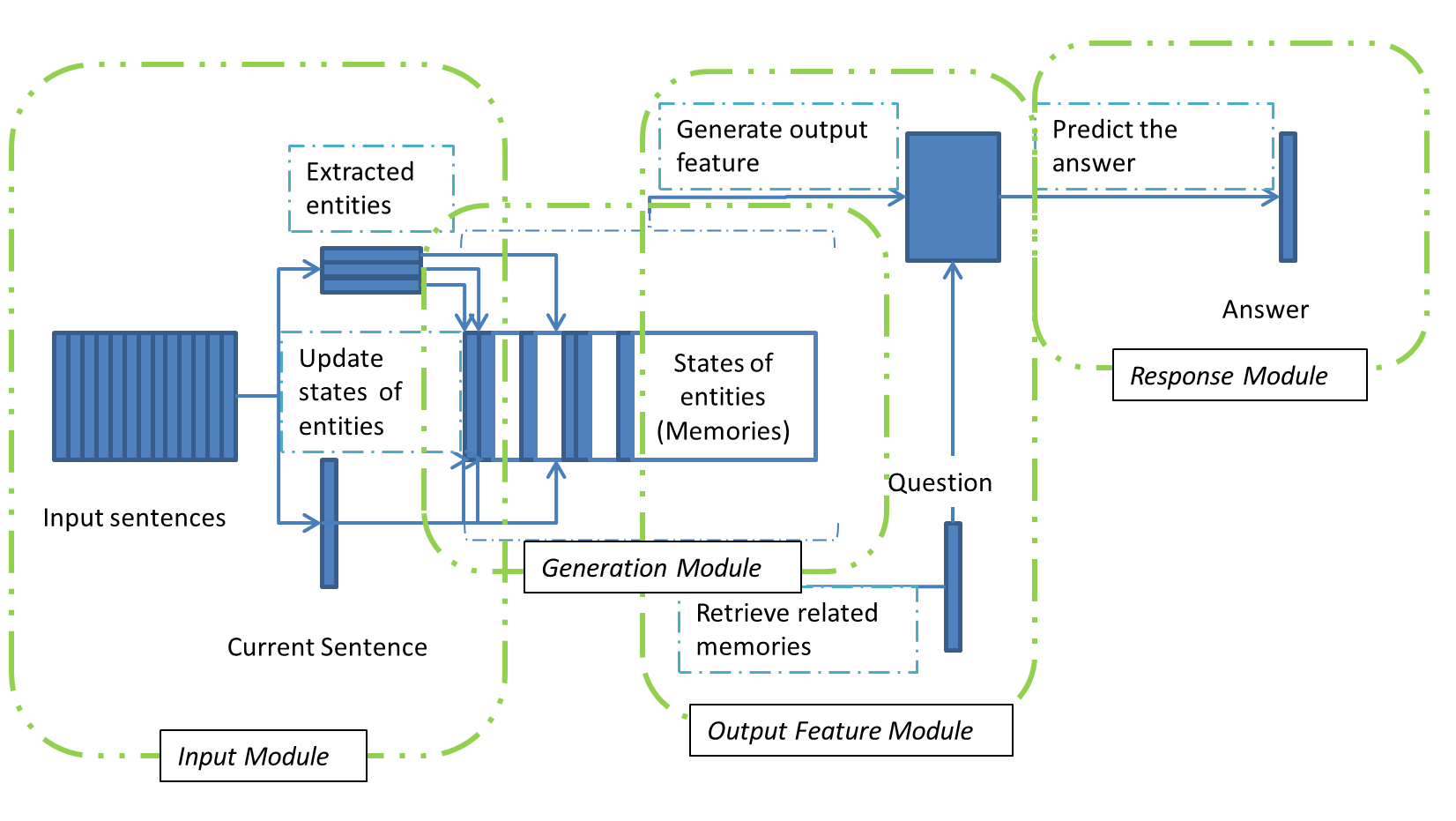}
  \caption{Architecture of the entity-based memory network.\\  The model is divided into four modules which are shown in the figure using squares.}\label{fig:arch}
\end{figure*}

The model monitors the entities involved in text and keeps updating their states according to the input. Whenever we have a question with regard to the text, we check the states of entities and predict an answer accordingly. The proposed model comprises of 4 modules, as is shown in Fig. \ref{fig:arch}.
Each module is designed for a unique purpose and together they construct the \emph{entity-based memory network} model.

\begin{enumerate}
\item I: Input module. Take as input a sentence and turn it into a vector. Meanwhile, extract all the entities it contains. The question is also processed using this module.
\item G: Generalization module. Update the states of related entities according to the input. For entities that are not contained in the memory pool, create a new memory slot for each of them and initialize these slots using pre-learned word embeddings.
\item O: Output feature module. It is triggered whenever a question arrives. Retrieve related entities according to the input question and then produce an output feature vector accordingly.
\item R: Response module. Generate the response according to the output feature vector.
\end{enumerate}

\subsection{Entity-based Memory Network Model}
Here we present a formal description of the proposed model.
Assume we have sentences $S_1,S_2,...S_n$ whose entities are annotated in advance as $e_{1},e_{2},...,e_{m}$. 

\paragraph{Input Module} 
We firstly turn each sentence $S_i$ into its vector representation:
\begin{equation} 
\vec{S_i}=f_1(S_i)
\end{equation}

\paragraph{Generalization Module} 

For a sentence ${S_i}$, we collect all the entities it contains $\{e_1^i,...,e_k^i,...,e_j^i\}$. These entities' states $\{\vec{e_k^i}\}$ are simultaneously updated according to $\vec{S_i}$ as follows: 

\begin{equation} 
\vec{\{e_k^i\}}=\arg\min_{\vec{\{e_k^i\}}} ~~(|\vec{S'_i}-\vec{S_i}|);\vec{S'_i}=f_2(e_1^i,...,e_k^i,...,e_j^i);   
\end{equation}

$f_2$ is to reconstruct $\vec{S_i}$ only using the states of $S_i$'s containing entities $\{\vec{e_k^i}\}$. $\{\vec{e_k^i}\}$ are updated to minimize the difference between $\vec{S'_i}$ and $\vec{S_i}$. Recall that $\vec{S_i}$ is generated using $f_1$ with the whole sentence $S_i$ as input. We compress the information carried by $S_i$ into a vector $\vec{S_i}$ and then unfold it into $\{\vec{e_k^i}\}$.

After processing these sentences, we construct a memory pool which consists of entities whose states are regarded as capable of representing the information carried by the input text.

\begin{figure*}[!ht]
  \centering
    \includegraphics[width=0.8\textwidth]{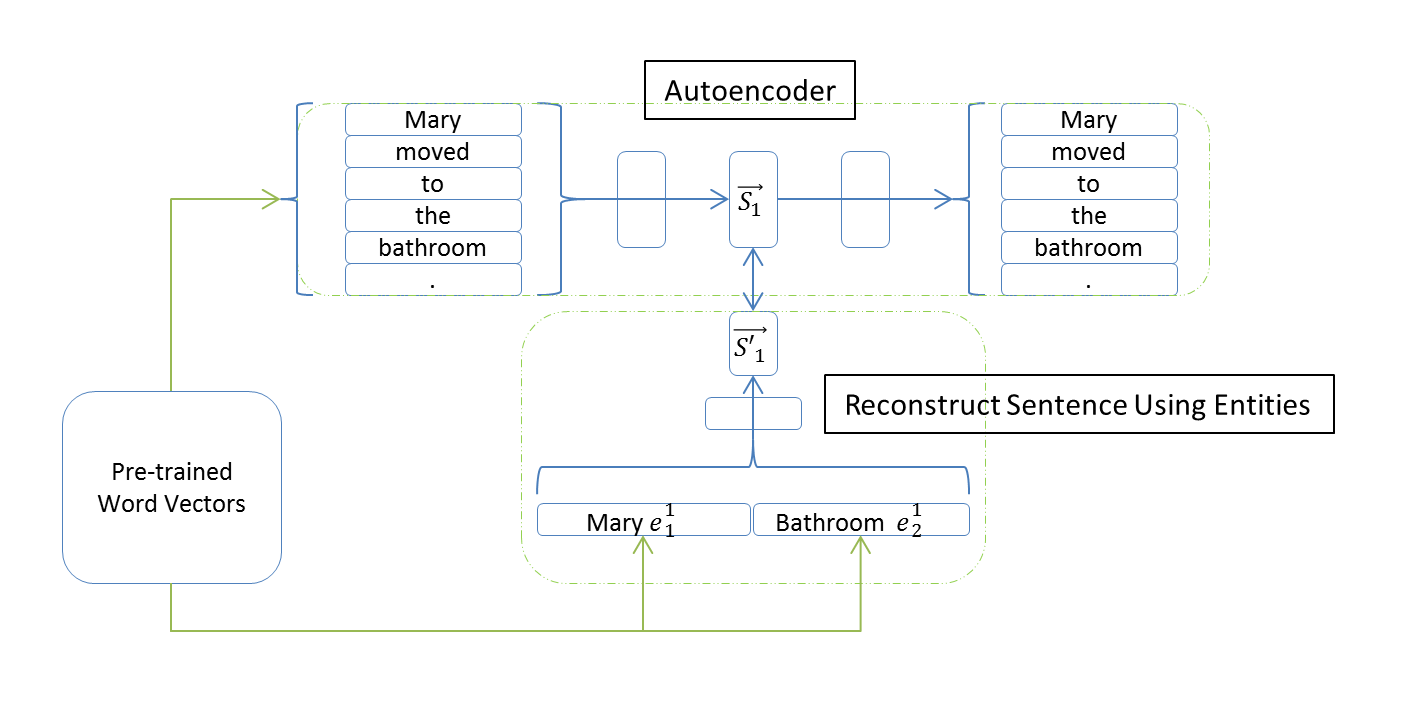}
  \caption{the Generalization Module: Using $S_1$ as an example, the autoencoder is used to convert the sentence into a vector $\vec{S_1}$ and the entities contained in $S_1$ are used to reconstruct the sentence vector.}\label{fig:aurnn}
\end{figure*}

\paragraph{Output Feature Module}
Question $q$ is turned into a vector $\vec{q}=f_1(q)$ and then $\vec{q}$ is used to retrieve related entities from the memory pool.

\begin{figure*}[!ht]
  \centering
    \includegraphics[width=\textwidth]{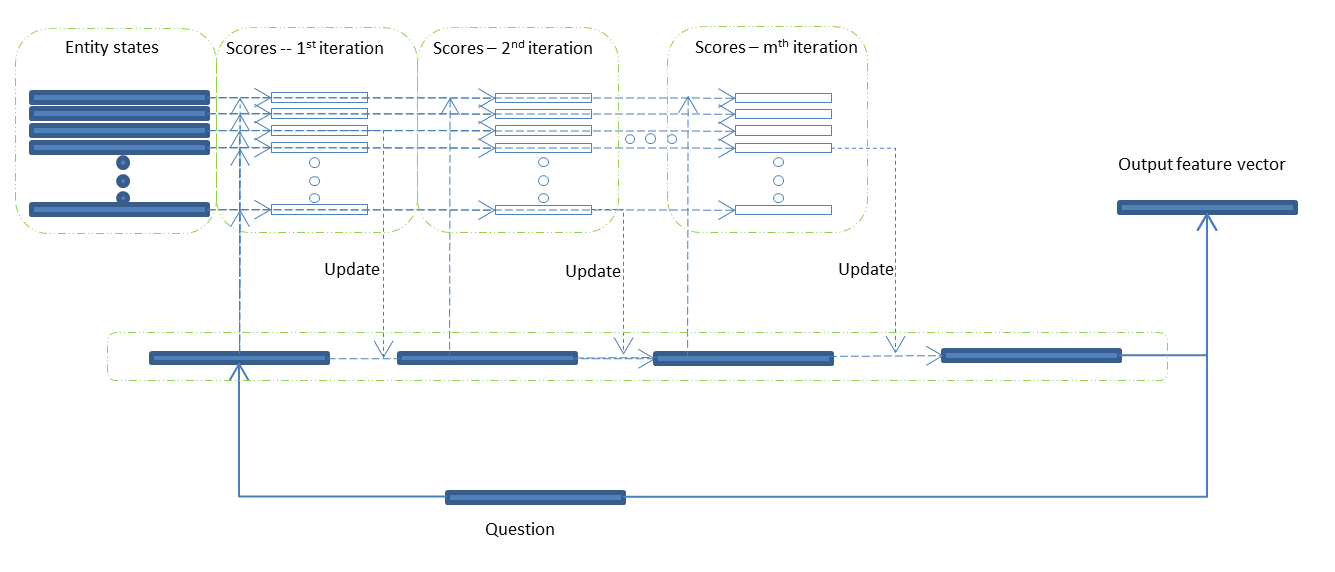}
  \caption{the Output Feature Module: In each iteration, entities are assigned different scores which indicate their importance in constructing the output feature vector.}\label{fig:res}
\end{figure*}

\begin{equation}
\left \{
\begin{aligned}
&\vec{O_0}=\vec{Q_0}=\vec{q},E_0=\phi\\
&\vec{Q_{j-1}}=h(\vec{Q_{j-2}},\vec{e_{j-1}}),j=2,3,...\\
&e_j=\arg\max_{e_k \notin E_{j-1}} p(\vec{e_k},\vec{Q_{j-1}});E_j=E_{j-1}\bigcup \{e_j\}\\
&\vec{O_j}=u(\vec{O_{j-1}},p(\vec{e_j},\vec{Q_{j-1}})*\vec{e_j})\\
\end{aligned}
\right .
\end{equation}

At first, $\vec{Q}$ in initialized using $q$. In the $j_{th}$ iteration, $p(\vec{e_k},\vec{Q_{j-1}})$ is the probability (or score) of $e_k$ being selected to compose the feature vector for answering $q$. Note that every $e$ is considered only once. In $\vec{Q}$, we consider the entity selected in the previous iteration. $\vec{Q}$ is kept updated using $\vec{e}$ and $p$. 

After several iterations, we use the final $\vec{O_m}$ as the output feature vector $\vec{O}$.
Note that if the $\vec{O_*}$ does not change much between iterations, we will omit the remaining loops. This early-stop strategy helps reduce the time cost.


\paragraph{Response Module}
Then we decide the answer using $a(q)=v(\vec{O})$. $a(q)$ produces a vector whose each item corresponds to one word in the vocabulary. $a(q)_i$ indicates the probability of $word_i$ being used as the correct answer. We choose the one with the highest probability.
Models like recurrent neural network can be used to output a sentence as the answer.

\subsection{Implementation}

This is a supervised model and requires annotated data for the training. The training data contains the input text, questions and answers. Also we need all the entities and entities that are related to the answer labeled.

We define the function form for training as follows:
As for $f_1$, many models, like the recurrent neural network, recursive neural network and so on \cite{mikolov2010recurrent,socher2011parsing,le2014distributed}, can be used to convert a sentence into a vector. 
Here we use an Long Short-Term Memory (LSTM) autoencoder \cite{li2015hierarchical} which takes a word sequence as input and outputs the same sequence.

$f_2$ takes a list of entity states as input and tries to reconstruct $\vec{S_i}$. We use the Gated Recurrent Unit (GRU) \cite{chung2014empirical}.
\begin{equation}
\begin{aligned}
\vec{S^{k}_i}&=tanh(GRU(\vec{S^{k-1}_i},\vec{e_k^i}))\\
\vec{S'_i}&=\vec{S^{j}_i}
\end{aligned}
\end{equation}

A GRU can be represented as the follows:
\begin{equation}
\left \{
\begin{aligned}
z_t^j&=\delta(W_z*x_t+U_z*h_{t-1})^j\\
\bar{h^j_t} &=tanh(W*x_t+U*(r_t\circ h_{t-1}))^j\\
r_t^j&=\delta(W_r*x_t+U_r*h_{t-1})\\
h_t^j&=(1-z_t^j)h_{t-1}^j+z_t^j\bar{h_t^j}\\
\end{aligned}
\right .
\end{equation}
$\circ$ represents an element-wise multiplication. $z_t^j$ and $r_t^j$ are two gates controlling the impact of historical $h_{t-1}^j$ on the current $h_t^j$. 
The GRU takes $x_t$ as input and updates the state of the neuron to $h_t^j$.
Compared with LSTM which it often replaces, it simplifies the computation while still keeps a memory of previous states. Therefore it takes less time to train GRU than LSTM.

Our goal is to minimize the loss $|\vec{S'_i}-\vec{S_i}|$. Using the stochastic gradient descent, we are able to train $f_2$ and also update $\{\vec{e_k^i}\}$. 
Note that the input module and the generalization module do not interact with the remaining. Thus they can be trained in advance.

The output feature module checks the memory pool repeatedly to select entities to form a feature vector:
\begin{equation}
\left \{
\begin{aligned}
&\vec{Q_{j-1}}=tanh(GRU(\vec{Q_{j-2}},\vec{e_{j-1}}))\\
&e_j=\arg\max p(\vec{e_j},\vec{Q_{j-1}})=\arg\max sigmoid(W*GRU(\vec{e_j},\vec{Q_{j-1}})+b)\\
&\vec{O_j}=tanh(GRU(\vec{O_{j-1}},p(\vec{e_j},\vec{Q_{j-1}})*\vec{e_j}))\\
\end{aligned}
\right .
\end{equation}


To generate the final answer, we use a simple neural network which takes the feature vector $\vec{O}$ as input and predict a word as output.
$p_w=v(\vec{O})=softmax(tanh(W'*\vec{O}+b))$. The word with the highest probability is selected. Suppose a sentence is to be generated, we  use the GRU to update $\vec{O}$ and then generate the sentence $\{w_*\}$ as follows:
\begin{equation}
\left \{
\begin{aligned}
\vec{p^{i-1}_w}&=softmax(tanh(W'*\vec{O_{i-1}}+b))\\
w_{i-1}&=\arg\max\vec{p^{i-1}_w}\\
\vec{O_i}&=tanh(GRU(\vec{O_{i-1}},\vec{w_{i-1}}))\\
\end{aligned}
\right .
\end{equation}

Similar to \cite{weston2014memory}, we use the stochastic gradient descent algorithm to minimize the loss function shown in Equation (\ref{eq:loss}) over parameters.
For an input $S_i$ and a given question $q$ annotated with the correct answer $word_a$ and related entities $\{e_{r}\}$, the loss function is as follows:

\begin{equation}\label{eq:loss}
\sum_{i\ne r} \max(0,\gamma - (p(e_r,q)-p(e_i,q)))+ \sum_{l\ne a} \max(0,\gamma -( p_{word_a}-p_{word_{l}}))+||\Theta||^2
\end{equation}

Here $\gamma$ is the margin and $||\Theta||^2$ is the squared sum of all parameters which is used for regularization.
Note that $\Theta$ does not include parameters of $f_1$ and $f_2$. Their parameters and states of entities are learned as described in Section 2.2.
Word vectors used to initialize entity states and words in autoencoder come from GloVe \cite{pennington2014glove}. The dimension is set to be 50.

\subsection{Data Preparation}
The model requires entities to be annotated in advance.
In this work, we treat each noun and pronoun as an entity. Different words are regarded as different entities for simplicity. This strategy saves us the effort of entity resolution which is a challenge for many languages. It also makes possible the application of the proposed model to entity resolution \footnote{We treat each mentions of entities as different one when processing the text and ask questions about which of these mentions refer to the same entities.}.
For datasets with related entities annotated, we can use the loss function described above. But annotating the related entities is time and labour-costing. Most datasets available are not annotated.
The weakly supervised learning can be applied to such data by trimming the loss function to  
\begin{equation}
\sum_{l\ne k} \max(0,\gamma -( p_{word_k}-p_{word_{l}}))+||\Theta||^2
\end{equation}

For unannotated data, a fully supervised training is also possible if we regard entities contained in questions as related entities or if we can use other methods to identify entities that are believed to be related.

\section{Experiments}
To verify the effectiveness of the proposed model, we conduct experiments on several datasets, including a toy QA data set bAbI \cite{weston2015towards}, the large movie review dataset for sentiment classification \cite{maas-EtAl:2011:ACL-HLT2011} and  the Machine Comprehension dataset (MC Test) \cite{richardson2013mctest}.
\subsection{bAbI}
The example shown in Fig. 1 is extracted from the bAbI dataset. It contains 20 topics, each of which contains short stories, simple questions with regard to the stories and answers. 
The data is generated with a simulation which behaves like a classic text adventure game. According to some pre-set rules, stories are generated in a controlled context.

Previous work reports extremely satisfying results using memory networks for most topics (around 90\% for most of them). However, we notice an interesting thing that all of them with no exception fail on the problem of path-finding which is to predict a simple path like "north, west" given the locations of several subjects. Another one is the positional reasoning. The Memory Network \cite{weston2015towards} reports accuracies of 36\% and 65\% for the two topics. The Dynamic Memory Network \cite{kumar2015ask} reports accuracies of 35\% and 60\%. The proposed model (Entity-MNN) reports accuracies of  53\% and 67\% respectively. It is still far from satisfying but the improvements on the two tasks indicates the superiority of the entity-based memory network. For the whole dataset, we report mean error rates about 12\%, comparable to 3.2 to about 24 reported by previous work \cite{sukhbaatar2015end,kumar2015ask,weston2015towards}.

The data is generated in a controlled text. As we know, QA systems trained on controlled text normally suffers when moving to real world problems \cite{hermann2015teaching}. Results on this toy dataset is not as convincing as that on practical tasks. 
Given how the bAbI data is generated, it is easy to achieve a 100\% accuracy if we do simple reverse engineering to identify the entities and rules. The good results of memory networks, including our model, can not be solely attributed to their ability of comprehension. It may be partly due to their ability of inducting the entities and rules from text.

\subsection{Machine Comprehension Test}
We tested the proposed model on a dataset constructed from children stories.
The machine comprehension test (MCTest) dataset \cite{richardson2013mctest} has 500 stories and 2000 questions (MC500). All of them are multiple choice reading comprehension questions. 
An additional smaller dataset with 160 stories and 640 questions (MC160) is also included in the MCtest data and used in our work.

Since the proposed model does not consider the form of multiple choice questions, we need to convert MCTest data into suitable formats firstly.
When answering a multiple choice question, one is provided with several alternatives of which at least one is correct.
These alternatives can be regarded as information known. 

For a question, we replace the ``Wh-" words using each alternative and Each alternative is turned to a new declarative sentences. These generated declarative sentences are generally understandable though may not be grammatically correct. Then we use the proposed system to decide whether the generated sentences are correct or wrong.  However, we do not distinguish between questions with only one answer and those with more than one answers as these newly generated sentences are treated separately. In other words, all questions are treated as having multiple answers.


The MCTest contains only hundreds of stories and is usually used for test only as statistical models normally require a large amount of training data. However, we still obtain satisfying results using this dataset. 
Table \ref{tab:MCtest} demonstrates the effectiveness of the entity-based model on the MCTest dataset. We outperform the previous state-of-the-art \cite{wang2015machine,sachan2016machine} on both MC160 and MC500. Our model does not employ rich semantic features as others do, and hence is easy to be migrated to languages aside from English.

\begin{table*}
\centering
\begin{tabular}{l | c| c| c| c| c| r}
  \hline
 Sys. & \multicolumn{3}{c|}{Acc.(\%) MC160} & \multicolumn{3}{c}{Acc.(\%) MC500}\\
  \hline
Type & Single &Multiple & Average & Single &Multiple & Average\\ \hline
  Richardson'13 \cite{richardson2013mctest}&76.8& 62.5& \cellcolor{gray}{69.2}& 68.0& 59.5& \cellcolor{gray}{63.3} \\ \hline
  Wang'15 \cite{wang2015machine}& 84.2 &67.9 & \cellcolor{gray}{75.3}& 72.1&67.9 &\cellcolor{gray}{69.9} \\ \hline
  Sachan'16 \cite{sachan2016machine}& - & - & -& 72.0&68.9 &\cellcolor{gray}{70.3} \\ \hline
\cellcolor{gray} {EntityMNN} & \multicolumn{3}{c|}{\cellcolor{gray} {Average=76.1}} & \multicolumn{3}{c}{\cellcolor{gray} {Average=76.6}} \\
  \hline
\end{tabular}
\caption{Results on Machine Comprehension Test}\label{tab:MCtest}
\end{table*}
\vspace{-1.1cm}

\subsection{Large Movie Review Dataset}
We further tested our model on the Large Movie Review Dataset \cite{maas-EtAl:2011:ACL-HLT2011}, which is a collection of 50,000 reviews from IMDB, about 30 reviews per movie. Each review is assigned a score from 1 (very negative) to 10 (very positive). The ratio of positive samples to negative samples is 50:50. Following the previous work \cite{maas-EtAl:2011:ACL-HLT2011}, we only consider polarized samples with scores no greater than 4 or no smaller than 7.

For each review, we present it as a short story and then add a question ``what is the opinion?". The answer is either ``negative" or ``positive". In this way we turn this task into a question answering problem. Note that although here the answer to a question is either ``negative" or ``positive", we do not put any constraints on the output. It is treated in the same way as open domain question answering and the system is expected to learn to predict the output by itself.

\begin{table}
\centering
\begin{tabular}{l |c | c|c|r}

  \hline
 Sys. &   Maas'11 \cite{maas-EtAl:2011:ACL-HLT2011} & Johnson'14 \cite{johnson2014effective}& Johnson'15 \cite{johnson2015semi}& \cellcolor{gray} {EntityMNN} \%\\
  \hline
Acc. & 89& 93.4 & 95 &\cellcolor{gray} {97.2}\\
  \hline
\end{tabular}

\caption{Results on Large Movie Dataset}\label{tab:movie}
\end{table}


We do not use the full dataset as the training takes a long time. We randomly select 10K samples (5K negative + 5K positive) for training and another 10K for test. We obtain an accuracy of 97.2\% on the subset which is higher than previous work \cite{maas-EtAl:2011:ACL-HLT2011,johnson2014effective,johnson2015semi} as is shown in Table \ref{tab:movie}. 
By exploring relations between entities, we consider information that is usually not included for classification tasks and obtain better results.

\subsection{Analysis}
The proposed model is designed based on the assumption that entities are the core of text. By updating the states of entities, information carried by text is encoded into entities. Thus all questions which are related to the text can be answered based on entities solely.

Using entities enable us to break a sentence into smaller text units and analyze text from a smaller scale. As stated, if one entity $e_i$ in sentence $S_a$ interacts with another entity $e_j$ in sentence$S_b$, dealing with $e_i$ and $e_j$ directly is much easier than dealing with $S_a$ and $S_b$. Therefore the proposed model overcomes this problem as has been proven in our experiments. 
A shortcoming with the proposed model is that, it cannot handle text that contains very few entities. Also hidden entities are not considered. As we know, pro-drop languages, like Japanese and Chinese, tend to omit certain classes of pronouns when they are inferable. The proposed model will encounter problems when dealing with such text.

\section{Conclusion}
This work presents the entity-based memory network model for text comprehension. All the information conveyed by text is encoded into the states of its containing entities and questions regarded to the text are answered using these entities. Experiments on several tasks have proven the effectiveness of the proposed model.
The proposed model is based on the assumption that entities can express all the information of text. In future research, we will further explore its ability by considering more components in text.

\bibliographystyle{splncs03}
\bibliography{emn}  
\end{document}